\pdfoutput=1

\documentclass[11pt]{article}

\usepackage[preprint]{acl}

\usepackage{times}
\usepackage{latexsym}
\usepackage{svg}

\usepackage[T1]{fontenc}

\usepackage[utf8]{inputenc}

\usepackage{microtype}

\usepackage{inconsolata}

\usepackage{graphicx}

\usepackage{enumitem}
\usepackage{booktabs}

\title{Understanding Language Model Circuits through Knowledge Editing}

\author{Huaizhi Ge \\
  Columbia University \\
  \texttt{hg2590@columbia.edu} \\\And
  Frank Rudzicz \\
  Dalhousie University \\
  \texttt{frank@dal.ca} \\\And
  Zining Zhu \\
  Stevens Institute of Technology \\
  \texttt{zzhu41@stevens.edu} \\}

\begin{document}
\maketitle
\begin{abstract}
Recent advances in language model interpretability have identified circuits, critical subnetworks that replicate model behaviors, yet how knowledge is structured within these crucial subnetworks remains opaque. To gain an understanding toward the knowledge in the circuits, we conduct systematic knowledge editing experiments on the circuits of the GPT-2 language model \citep{radford2019language}. Our analysis reveals intriguing patterns in how circuits respond to editing attempts, the extent of knowledge distribution across network components, and the architectural composition of knowledge-bearing circuits. These findings offer insights into the complex relationship between model circuits and knowledge representation, deepening the understanding of how information is organized within language models. Our findings offer novel insights into the ``meanings'' of the circuits, and introduce directions for further interpretability and safety research of language models.
\end{abstract}

\section{Introduction}
The interpretability of large language models (LLMs) has advanced significantly with the exploration of ``circuits''—specific subsets of model parameters tailored for distinct tasks. While these circuits, derived from specialized training on text datasets, have proven crucial for replicating model behaviors, a question has not been studied thoroughly: How is knowledge structured within these critical subnetworks? Previous works have approached circuit analysis through ablation studies of designated components \citep{conmy2023towards}. Others have attempted to verbalize network components, either using the language model head as a ``lens'' \citep{nostalgebraist2020Interpreting,belrose2023Eliciting} or an external model \citep{hernandez2022Natural,singh2023Explaining}, but these approaches are expensive to scale to the study of the knowledge representation properties of the automatically-extracted circuits.

We propose a novel perspective to understand the knowledge structure within automatically-extracted circuits through systematic knowledge editing experiments. This approach examines circuits as holistic objects rather than analyzing individual components, offering insights into how information is organized and modified within these crucial subnetworks.

Using the GPT-2 base model \citep{radford2019language}, we first investigate knowledge storage patterns within circuits derived from diverse text classification datasets. These circuits were adapted to generation tasks to facilitate knowledge editing experiments. Through knowledge editing, we assess how different circuit components respond to knowledge modifications. Our findings reveal an intriguing ``confirmation bias'' behavior: Knowledge-intensive circuits demonstrate stronger resistance to editing compared to their complementary parts, suggesting structured information storage patterns.

To better understand knowledge distribution, we explore circuits of varying sizes, from 50\% down to 5\% of the model's parameters. This investigation reveals that the ideal knowledge-bearing circuit likely lies between these bounds, indicating that knowledge is neither extremely concentrated nor widely dispersed throughout the network.

Our analysis of cross-dataset circuit overlap provides further insights into knowledge organization. Using mask similarity analysis, we find nuanced patterns of overlap between circuits from different tasks, particularly between linguistic and knowledge-based datasets. 

Examining the architectural composition of knowledge-bearing circuits, we make a surprising discovery: LayerNorm components constitute a larger share of the circuits than traditionally emphasized attention and multi-layer perceptron (MLP) layers. This finding provides novel evidence that complements conventional assumptions about knowledge storage in neural networks and highlights the potential importance of normalization layers in maintaining network stability and information organization.

These findings collectively enhance our understanding of how neural networks structure and modify information within circuits. By revealing patterns in knowledge distribution, circuit overlap, and architectural composition, our study introduces new directions for interpretability research and potential implications for model safety. Our results suggest that effective manipulation of model behavior may require careful consideration of how knowledge is organized within these crucial subnetworks. All our codes and analysis data is open-sourced at GitHub.


\section{Related work}
\paragraph{Circuit analysis} Circuit extraction tries to find a minimal subnetwork that represents the behavior of the full network (computation graph), where the components can be explainable \citep{olah2020zoom,elhage2021mathematical}. This extracted circuit should be faithful \cite{hanna2024have}: if the non-circuit graph edges outside of the circuit are removed, the model's performance should stay the same. Each component of the circuit (the vertex of the graph) is usually a module \cite{conmy2023towards,wang2022interpretability} or a causal variable that supports interventions \cite{vig2020investigating,geiger2021causal}. 

While the circuit analysis literature study the module level, alternative granularity levels exist, analyzing on parameter level \citep{csordas2020neural,bayazit2023Discovering} and sparse autoencoder level \citep{marks_sparse_2024}. In this paper, we focus on the parameter level, but our approach can be extended to other granularity levels without loss of generality.

A contemporaneous work, Knowledge Circuits \citep{yao_knowledge_2024}, also studies the circuit from a knowledge editing point of view. Our works differ in the granularity level of the circuit, the types of knowledge, the tasks studied, and the circuit editing methods.

\paragraph{Model editing} Various methods modify knowledge in large language models. \citet{zhu2020modifying} introduced a task for selective updates of factual knowledge in Transformer models, using fine-tuning as a benchmark. Rank-One Model Editing (ROME) by \citet{meng2022locating} adjusted feed-forward weights to change specific factual associations. MEMIT \citep{meng2022memit} enabled integrating multiple memories into a language model. Model Editor Networks with Gradient Decomposition (MEND) by \citet{mitchell2021fast} utilized a single targeted input-output pair for rapid, localized changes in a pre-trained model’s behavior. Other influential approaches include targeting specific knowledge neurons \citep{dai2021knowledge} and using hyper-networks \citep{de2021editing}. These techniques have proven to be highly effective in editing knowledge within LLMs.

\section{Methods}
\paragraph{Circuit extraction}
We adopt a differentiable masking technique implemented in \citet{circuit_detection_2024}.  
Similar to the differentiable masking literature \citep{de-cao-etal-2022-sparse,csordas2020neural,bayazit2023Discovering}, we set up learning objectives to train a mask parameter $m_\theta \in [0,1]$ for each model parameter $\theta \in \Theta$, where $\Theta$ describes the collection of all trainable parameters in the full model. Unlike \citet{de-cao-etal-2022-sparse}, we use a combination of the following objectives:
\begin{enumerate}
    \item Faithfulness loss $L_F=\frac{1}{N} \sum_i \sum_{j}\hat{y}^{(i)} \log\frac{1}{\hat{l}_j^{(i)}}$, which is a cross-entropy loss. This objective encourages $\hat{l}_j^{(i)}$, the logits of the circuit (i.e., the masked model) at class $j$ of sample $i$, to predict the same results as the original model $\hat{y}^{(i)}$. This loss is averaged across the $N$ training data samples.
    \item Sparseness loss $L_S=\frac{1}{|\theta|}\sum_{\theta} \sigma(m_\theta)$, where $m_\theta$ is the mask for the model parameter $\theta$. This approximately counts the density of the nonzero masks (with a sigmoid function $\sigma(\cdot)$ that smooths the distribution). This encourages the circuit to be sparse.
\end{enumerate} 
Following \citet{de-cao-etal-2022-sparse}, we use the hard-concrete distribution of \citet{louizos2018learning} to sparsify the masks. Following \citet{circuit_detection_2024}, we use a straight-through estimator \citep{bengio2013estimating} to make the forward pass (apply the mask) trainable. We train until the sparseness reaches a pre-defined threshold. 

\paragraph{Circuit-aware model edit}
In the circuit extraction step, we identified a circuit that is a subset of the model. To study how circuits are affected by the model's knowledge, we replace the classification head of the model with its language model head, transforming it into a generation model. Thus, we repurpose the identified circuits from text classification tasks for generation tasks, allowing us to conduct knowledge editing. 

This circuit-aware model edit involves altering the true target to a new target. For instance, given the prompt ``A cat is a kind of'', we would edit the true target ``animal'' to the new target ``plant''. Throughout this study, we employ fine-tuning techniques to modify the knowledge stored within these circuits.

The circuit-aware model edit is implemented using the fine-tuning method from the EasyEdit repository \citep{wang2023easyedit}, where we modify the knowledge embedded in the model with a cross-entropy loss:
\begin{equation}L_C = \frac{1}{N} \sum_{i} \sum_j y^{(i)} \log\frac{1}{p_j^{(i)}},
\label{eq:circuit-aware-cross-entropy-loss}
\end{equation}
where $N$ is the number of samples, $y^{(i)}$ is the true label for the $i$-th sample (0 or 1), and $p_j^{(i)}$ is the predicted probability of the $i$-th sample being in class $j$. We use Stochastic Gradient Descent (SGD) to update the parameters:
\begin{equation}\theta_{t+1} = \theta_t - \eta m_\Theta \nabla_{\theta} L_C(\theta_t; x^{(i)}, y^{(i)}),
\label{eq:model-edit-updating-equation}
\end{equation}
where $\theta_t$ are the parameters at iteration $t$, $\eta$ is the learning rate, and $m_\Theta$ is the collection of all binary parameter masks $m_\theta$ and is kept constant during the SGD. For parameters involved in the circuit $\mathcal{C} \subseteq \mathcal{P}$, the mask is set to 1; otherwise, it is set to 0. The term $\nabla_{\theta} L_C(\theta_t; x^{(i)}, y^{(i)})$ represents the gradient of the loss function $L_C$ with respect to the parameters $\theta_t$, computed using a single training example $(x^{(i)}, y^{(i)})$ or a mini-batch of examples.

In this study, we specifically focus on editing the circuits within the model. Following the differential masking literature \citep{bayazit2023Discovering,circuit_detection_2024}, the parameters not involved in these circuits are masked by zero and are maintained at zero throughout the training process by setting their gradients to zero in each step, ensuring they do not affect the model's behavior.

\begin{figure*}
  \centering
    \includegraphics[width=\columnwidth]{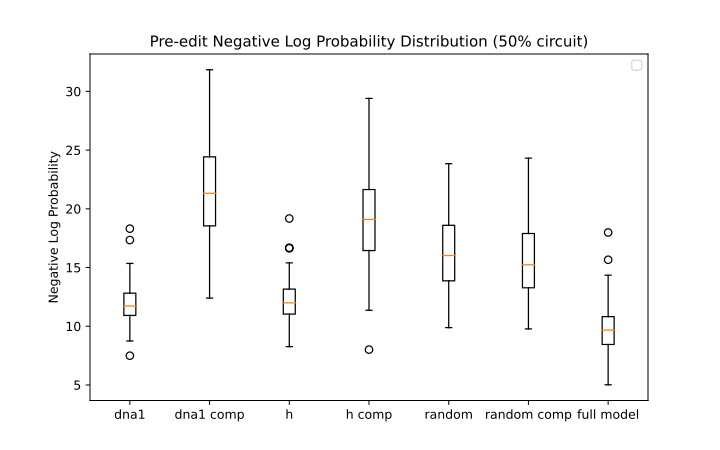}
    \includegraphics[width=\columnwidth]{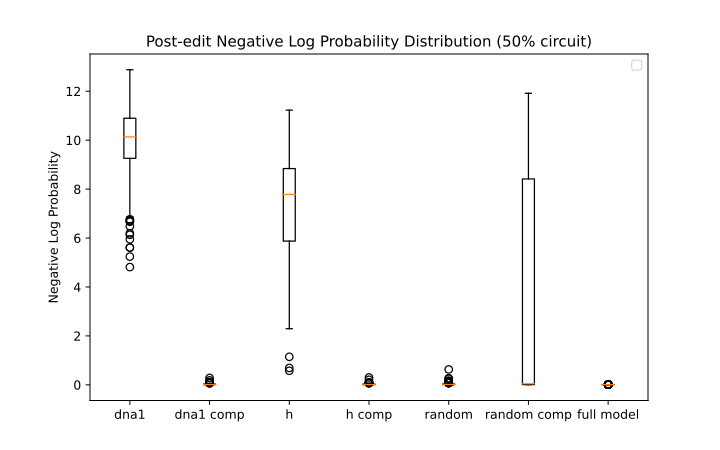}
  \caption{Negative log probability distribution of the 50\% circuits. Left: distribution of the true targets pre-edit. Right: distribution of the new targets post-edit.}
  \label{part1}
\end{figure*}

\section{Data}
\subsection{Model edit data}
We use \textsc{Hierarchy Data} \citep{perplexing_2024} to act as the model edit data. This dataset encompasses a series of both challenging incorrect facts, represented as $(s, r, o^*)$, and their corresponding accurate facts, denoted as $(s, r, o)$. Here, $s$ signifies the subject and $o$ the object, both selected from the hierarchy chains. The relation $r$ consistently adopts the hypernym schema, emphasizing hierarchical connections. An example of this is ``A British Shorthair is a kind of cat''. We modify the objects of these facts, generating altered facts ``A British Shorthair is a kind of dog'' to test the efficacy of circuit edits. The \textsc{Hierarchy Data} dataset includes 198 facts targeted for editing analysis. This structured approach facilitates explorations into the role of hierarchical relations in the adaptability and accuracy of language model editing processes. 

\subsection{Text classification data}
We include a range of datasets spanning hierarchical concept knowledge, linguistic knowledge, and those involving model's safety. Each dataset is re-formatted into a classification task that facilitates circuit extraction.

\paragraph{Hierarchical data}
Here we re-format \textsc{Hierarchy Data}(h) into a classification dataset. For instance, we categorized the statement ``A cat is a kind of animal'' with a label of 1, indicating a true assertion, whereas the statement ``A cat is a kind of plant'' was labeled as 0, denoting a false assertion.

\paragraph{Linguistic knowledge} We include eight datasets from BLiMP \citep{warstadt2020blimp}. These cover both the syntax and semantic aspects, allowing us to examine if the edits affect the linguistic knowledge of the language models. The datasets are:
\begin{itemize}[nosep]
    \item \texttt{determiner noun agreement 1} (dna1)
    \item \texttt{determiner noun agreement 2} (dna2)
    \item \texttt{determiner noun agreement irregular 1} (dnai1)
    \item \texttt{determiner noun agreement irregular 2} (dnai2)
    \item \texttt{determiner noun agreement with adjective 1} (dnawa1)
    \item \texttt{determiner noun agreement with adj 2} (dnawa2)
    \item \texttt{determiner noun agreement with adj irregular 1} (dnawai1)
    \item \texttt{determiner noun agreement with adj irregular 2} (dnawai2)
\end{itemize}

\paragraph{Safety behavior}
A crucial goal for the interpretability analysis of models is to make the models safer. To this end, we use five datasets from \citet{perez2022discovering} that describe the safety-related behavior of language models in the real world, including:
\begin{itemize}[nosep]
    \item \texttt{risk averse} (ra)
    \item \texttt{extraversion} (e)
    \item \texttt{desire for being rated helpful, harmless \& honest over actually being helpful, harmless \& honest} (hhh)
    \item \texttt{okay with using many resources} (umr)
    \item \texttt{desire for advancing technology to achieve goals} (tech)
\end{itemize}

\section{Validating the circuit extraction}

To validate the accuracy and correctness of the extracted circuits, we evaluate their performance across specific classification tasks. In particular, we focus on two types of circuits: those containing 5\% parameters and those containing 50\% parameters. 

Table~\ref{circuitaccuracy} presents the task accuracy achieved by these identified circuits. The circuits were evaluated based on their ability to reproduce task outcomes. High accuracy scores indicate that the circuits not only preserve the task-relevant information but also reflect the underlying computational mechanisms of the original model.
The results show that both the 5\% and 50\% circuit subsets achieve high levels of accuracy, suggesting that the circuits have been correctly identified.

\begin{table*}
  \centering
  \begin{tabular}{llllll}
    \toprule
         & \textbf{h circuit}     & \textbf{dna1 circuit} & \textbf{dna2 circuit}  & \textbf{ra circuit} & \textbf{e circuit} \\
    \midrule
    \textbf{5\% circuit}     & 0.84 & 0.91  & 0.87  & 0.94 & 0.97  \\
    \textbf{50\% circuit}     & 0.88  & 0.89  & 0.87  & 0.94 & 0.87 \\
    \bottomrule
  \end{tabular}
  \caption{\label{circuitaccuracy}Task accuracy of identified circuits}
\end{table*}

\section{How knowledgeable are the LLM circuits?}
\label{sec:circuits-knowledgeable}
\subsection{Experiments}
To investigate the amount of knowledge in the LLM circuits, we edit the knowledge within a specific circuit and compare its performance with that of its complementary circuit. Typically, individual circuits represent only a small percentage of the model's total parameters. However, for a clearer demonstration of the amount of knowledge in the circuits, we analyze the 50\% circuits, ensuring that these had an equivalent number of parameters as their complementary circuits. 

In addition, we also refer to the remainder part of the model's parameters ``the complementary circuit''. For example, the complementary circuit of the dna1 task is referred to as ``dna1 complementary'' (dna1 comp for short).

\subsection{The circuit vs complementary circuit}
Even before applying any edits, we want to understand how the circuit and the complementary circuit encode entity-related knowledge, so we query the pre-edit negative log probabilities of the true targets. A lower values indicate a stronger grasp of the knowledge. 

Figure~\ref{part1} (left) displays the results. The extracted circuits demonstrate significant lower ($p<0.05$) negative log probabilities of the \textsc{HierarchyData} entities compared to their complementary circuits, indicating a superior understanding of the knowledge. Note that none of the circuits outperforms the full model, showing that the complementary circuits still capture some factual knowledge.

A noteworthy observation is that this effect generalizes across both the h circuit (which is derived from the same data as the \textsc{HierarchyData} entities) and the dna1 circuit (which is tasked with a distinct, syntax-focused dataset). This finding suggests that the model's circuits for solving the syntax tasks may be relevant to the model's mechanism for storing knowledge, thus providing a potential explanation for the ``spill-over'' effects \citep{sahak-etal-2023-state} --- i.e., the effects of some datasets on the models may ``spill over'' to other apparently irrelevant datasets. We will study the overlap of circuits between different tasks in Section \ref{sec:circuits-overlap}. 

As a baseline, the probability of a random circuit is not significantly different from that of its complementary circuit, confirming that the aforementioned ``circuit vs complementary'' differences do not arise from randomness. 

\subsection{The complementary circuit is more susceptible to the model editing}
After applying the circuit-aware knowledge editing, we query the negative log probabilities of the \textsc{HierarchyData} entities again, and the outcomes are presented in Figure~\ref{part1} (right).


Our metric for assessing edit performance is the negative log probability of the new target, where lower values indicate better editing. The figure displays results for circuits extracted using the ``determiner noun agreement 1'' (dna1) dataset of BLiMP, circuits extracted using hierarchy data (h) along with their complementary circuits, a randomly extracted circuit alongside its complementary circuit, and the full model circuit. Further details about other datasets are provided in Appendix \ref{app:post_vs_pre_edit_50}. Notably, the complementary 50\% circuit consistently outperforms the 50\% circuit in terms of editing for both the dna1 and h circuits. Furthermore, the performance of the random circuit is similar to that of its complementary counterpart. These indicate that the complementary circuits, which are less task-relevant, are more susceptible to knowledge edits. Yet, the full model demonstrates the best editing performance, indicating that the circuit might still synergize with the complementary networks upon the injection of new knowledge.

Similar to the pre-edit observations, the post-edit discrepancies between the circuits and their complements can be generalized to a linguistic task (dna1). This suggests that task-specific circuits might indicate the knowledge storage locations. In other words, altering the knowledge from these knowledge-rich circuits is more difficult than altering the knowledge from their complements. 

Interestingly, humans have similar ``confirmation bias'' behavior, tending to reject hypotheses that do not agree with prior assumptions \citep{wason1960failure}. Our findings here can also provide a novel perspective about neural network plasticity \citep{lyle2023Understanding}, but a further in-depth inquiry into the nature of circuit-specific plasticity is left to future works.

\begin{figure*}[t]
  \centering
  \includegraphics[width=\columnwidth]{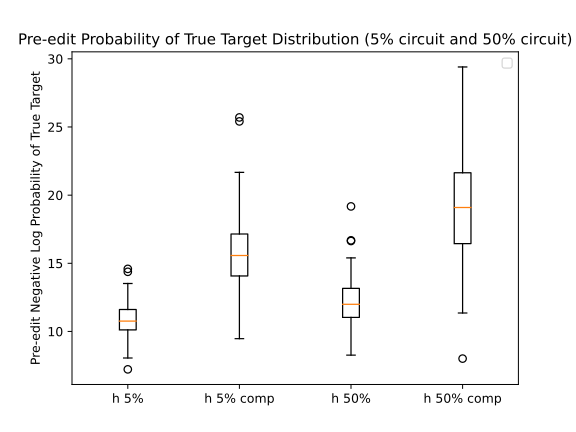}
  \includegraphics[width=\columnwidth]{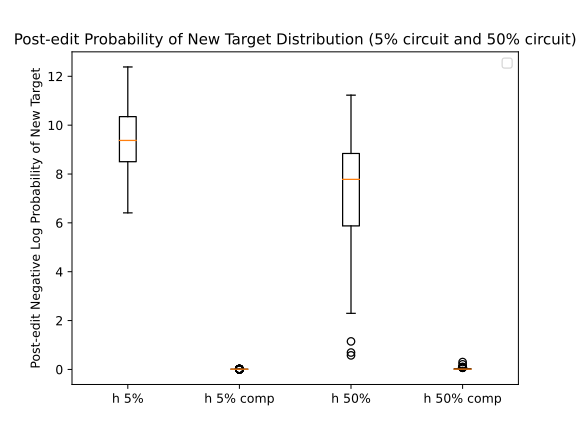}
  \caption{Pre-edit (left) and post-edit (right) log-probability distribution of 5\% circuit and 50\% circuit}
  \label{part2}
\end{figure*}

\section{How large are ``ideal circuits''?}
\label{sec:circuit-size-reduction}
Suppose that there exists a circuit that stores all knowledge needed to perform the task --- let's use the term ``ideal circuit'' to describe it --- how large would it be?

In our previous experiments, we conducted experiments on circuits that comprised 50\% of the total model parameters. To investigate potential disparities, we reduced our analysis to include circuits that accounted for 5\% of the parameters, using \textsc{HierarchyData}. 

Figure~\ref{part2} (left) presents the negative log probability scores of the true targets before model edits, and Figure~\ref{part2} (right) illustrates the outcomes of editing the 5\% vs. the 50\% circuits, along with their corresponding complementary circuits. We continue to employ the negative log probability of the new targets to quantify the edit performance. Detailed figures of 5\% circuits derived from other datasets are presented in Appendix \ref{app:post_vs_pre_edit_5}.

Figure~\ref{part2} shows that both the 5\% and the 50\% circuits are harder to edit than their corresponding complementary circuits. Notably, the 5\% circuit displays a higher level of such ``confirmation bias'' than the 50\% circuit. This suggests that the 5\% circuit may be closer to an ideal circuit, than the 50\% circuit.

Now, if an ideal circuit exists, would its size be smaller than 5\% or between 5\% and 50\%? 

As the pre-edit logprob distribution plot in Figure~\ref{part2} (left) shows, the 5\% circuit captures the knowledge the best, followed by the 50\% circuit (perhaps because the 50\% circuit may contain non-knowledgeable ``noise'' parameters that mask out the circuit's ability to demonstrate knowledge). Also shown in this figure, the complementary of 5\% circuit is more knowledgeable than the complementary of 50\% circuit, indicating that the 5\% complementary circuit likely possesses certain key parameters directly associated with knowledge retention, whereas the 50\% complementary circuit may include a surplus of less critical parameters. In other words, an ideal circuit, if it exists, would incorporate more than 5\% but less than 50\% of parameters.

Table~\ref{percentages} provides additional evidence toward the sizes of the idealized circuits. We replicate the experiments, this time extracting 15\%, 25\%, 35\% parameters circuits while using the same hyperparameters. The table below shows the median post-edit negative log probabilities of the new target, combined with the results from the 50\% and 5\% circuits (higher values indicate worse performance). As expected, the edit performance of the extracted circuits declines as the circuit size decreases. If an ideal circuit exists that encapsulates all relevant knowledge in the model, it would likely have the worst edit performance due to strong ``confirmation bias''. Therefore, we hypothesize that the 5\% circuit may be closer to this ``ideal circuit'' than the 50\% circuit. However, the exact size of this idealized circuit remains to be further investigated.

\begin{table*}
  \centering
  \begin{tabular}{llllll}
    \toprule 
         & \textbf{50\% circuit}     & \textbf{35\% circuit} & \textbf{25\% circuit}  & \textbf{15\% circuit} & \textbf{5\% circuit} \\
    \midrule 
    \textbf{Circuit}     & 7.78 & 8.10  & 8.28  & 9.30 & 9.37  \\
    \textbf{Complementary circuit}     & 1.26e-2  & 9.99e-2  & 9.97e-2  & 9.85e-2 & 9.81e-2 \\
    \bottomrule
  \end{tabular}
  \caption{\label{percentages}Median post-edit negative log probabilities of the new target across circuits with different sizes}
\end{table*}

Recent papers on circuits detected highly sparse circuits that usually contain less than 1\% of the total parameters \citep{bayazit2023Discovering} or less than 3\% of total edges \citep{hanna2024have} for replicating the model behavior. We hypothesize that while the circuits less than 1\% explain the functions, the ``ideal circuits'' need more parameters that correspond to the knowledge used in the tasks. Also, considering that the knowledge localization might differ from the model editing \citep{hase2024does}, the roles of distinct parameters in the model would be an intriguing path to explore further.

\section{Do these circuits overlap?}
\label{sec:circuits-overlap}
The datasets used for circuit extraction in the previous sections showcase considerable diversity, yet they yield surprisingly consistent results. This consistency raises an interesting question: do these circuits significantly overlap?

To create a baseline, we randomly select a 5\% parameter circuit mask and calculate the overlap percentage of all parameters within the circuit. As shown in Table~\ref{random_overlap}, the percentages constantly appear around 5\% and only deviate by errors that are two magnitudes smaller. Mathematically, the expectation of a circuit with X\% parameter overlaps with a random X\% circuit is X\%.

\begin{figure}
  \centering
    \includegraphics[width=\columnwidth]{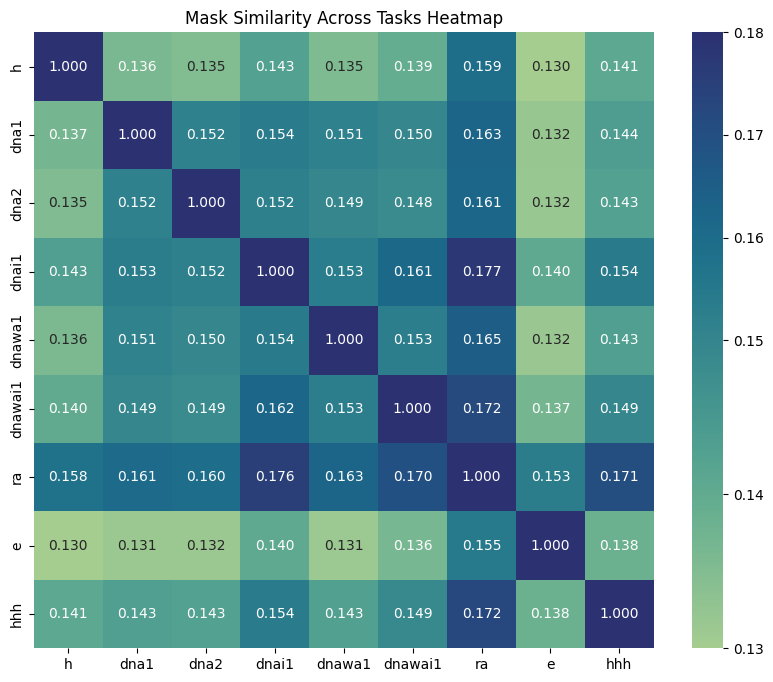}
  \caption{Mask similarity across tasks heatmap. The overlaps are neither too large nor too small.}
  \label{part3}
\end{figure}

Figure~\ref{part3} features a heatmap that depicts the mask similarity across different tasks. Each entry in the heatmap represents the degree of overlap between 5\% circuits from two datasets, measured relative to the base dataset listed in the row. For example, a value of 0.137 at the intersection of the second row and the first column suggests that the circuit derived from the dna1 dataset overlaps with the h dataset circuit by 13.7\% of the dna1 circuit.

\begin{table*}
  \centering
  \resizebox{\linewidth}{!}{
  \begin{tabular}{l lllllllll}
    \toprule 
         \textbf{Task} & \textbf{dna1} & \textbf{dna2}     & \textbf{h} & \textbf{dnai1}  & \textbf{dnawa1} & \textbf{dnawai1} & \textbf{e} & \textbf{ra} & \textbf{hhh} \\
    \midrule 
         \textbf{Difference from 0.05} & 2.6e-4 & 2.4e-4 & 1.3e-4 & 0.7e-4 & 2.1e-4 & 2.3e-4 & 0.1e-4 & -0.3e-4 & 0.7e-4 \\
    \bottomrule
  \end{tabular}}
  \caption{\label{random_overlap}Percentage of parameter overlap between each 5\% circuit and a randomly selected circuit}
\end{table*}

The degrees of overlap between the circuits vary across different dataset comparisons --- between 13\% and 18\% --- which are at least twice the baseline of 5\%. These indicate non-negligible similarities. Interestingly, datasets that appear closely related, such as dna1 and dna2, sometimes display significant differences when viewed through the lens of the model’s mechanism. Notably, the ra (risk-averse) circuit shows greater overlaps with other circuits, suggesting that the risk-averse dataset may include terms that are broadly relevant across multiple contexts. Conversely, the e (extraversion) circuit demonstrates fewer overlaps, implying that the extraversion dataset likely features specialized terms that do not commonly occur in other datasets.

Similarly, the 50\% circuit of three datasets overlap with each other at levels significantly higher than the 50\% expectation, as shown in Table~\ref{50overlap}.

\begin{table*}
  \centering
  \begin{tabular}{llll}
    \toprule 
         & \textbf{h circuit}     & \textbf{dna1 circuit} & \textbf{ra circuit}  \\
    \midrule 
    \textbf{h circuit}     & 1 & 0.61  & 0.58   \\
    \textbf{dna1 circuit}     & 0.61  & 1  & 0.59  \\
    \textbf{ra circuit}     & 0.58  & 0.59  & 1  \\
    \bottomrule
  \end{tabular}
  \caption{\label{50overlap}Percentage of parameter overlap between 50\% parameter circuits}
\end{table*}

Our finding echoes those of \citet{merullo2023Circuit} that the circuits of different tasks may overlap with each other. Our analysis also underscores nuanced relationships between different datasets and highlights how specific dataset characteristics might influence the generalizability of the circuits derived from them.

\section{What Transformer modules are in the circuits?}
In this section, we identify which types of layers are particularly influential within the extracted circuit. Table~\ref{part4} presents a detailed breakdown of the proportion of total parameters that each type of layer accounts for within the circuit. The data reveal that LayerNorm constitutes a large percentage of the circuit's parameters, especially noticeable within the 5\% circuit subset. In contrast, the attentions and multi-layer perceptrons (MLPs) account for a smaller, yet comparable, share of the circuits.

This distribution of parameters is intriguing because it suggests that, while LayerNorm plays a crucial structural role, the computational responsibilities might be more evenly distributed or perhaps differently focused in attention and MLP layers. Notably, some studies found that the MLPs are critical repositories of knowledge within neural networks \citep{geva2020transformer,geva-etal-2022-transformer}, and the attentions are critical for the mechanisms related to factual recall \citep{geva2023dissecting}. Recently, \citet{katz-belinkov-2023-visit} found that LayerNorm is related to the introduction of new tokens. There are also some recent debates about the exact localization of the knowledge \citep{hase2024does,niu2024does}. Complementing prior works, our observations show that LayerNorm plays a critical role in the circuits related to the syntax and hierarchical knowledge entity tasks.

The significant allocation of parameters to LayerNorm in the circuits raises questions about their specific function and impact on the overall network efficacy. Future analysis of LayerNorm could further advance our understanding.

\begin{table*}
  \centering
  \begin{tabular}{lllll}
    \toprule
         & \textbf{dna1 5\% circuit}     & \textbf{dna1 50\% circuit} & \textbf{h 5\% circuit}  & \textbf{h 50\% circuit} \\
    \midrule
    \textbf{attn} & 0.04  & 0.30  & 0.04  & 0.47   \\
    \textbf{mlp}     & 0.05 & 0.30  & 0.05  & 0.50    \\
    \textbf{ln}     & 0.37  & 0.63  & 0.37  & 0.62 \\
    \bottomrule 
  \end{tabular}
  \caption{\label{part4}Proportion of total parameters accounted for by each layer type within the circuit}
\end{table*}

\section{Discussion}
\paragraph{Safety-related model behavior is relevant to the factual knowledge stored in the LLMs} The relationship between safety-related model behavior and the factual knowledge stored in large language models (LLMs) is intricate and significant. During the circuit extraction phase, we utilized model behavior datasets, such as those indicating 'risk aversion' and 'extraversion', in Model-Written Evaluations. These behavior-related circuits appear to possess a substantial amount of factual knowledge. Understanding LLM safety fundamentally involves comprehending model behavior. To achieve this, it is essential to understand the knowledge embedded within the model. However, the precise nature of the knowledge or concepts required for models to determine their behaviors remains an open question. Additionally, the mechanisms by which models use these concepts to guide their behavior are not fully understood. These areas represent critical open problems in the field.

\paragraph{Remove biases without compromising the model's performance} One intriguing finding is that complementary circuits are easier to edit than the primary behavior-related circuits we identified. This suggests the possibility of performing editing tasks on complementary circuits without altering the primary circuits, thereby preserving the model's behavior. If validated, this approach could offer a method to remove biases through editing tasks without compromising the model's performance by using circuit identification.

\paragraph{Implications for model training} The understanding of the information structures in the circuits can lead to fine-tuning technologies. The neural network circuit is a particularly appropriate medium because the sparsity can lead to high efficiency, and the automatic circuit discovery algorithms allow the search of circuits that sacrifice model performance to the fewest possible extent. We plan to continue developing the training algorithms in future works.

\paragraph{Recommendations for safer models}
We propose several recommendations for future researchers aiming to enhance model safety. First, further investigation into the role and characteristics of complementary circuits can provide insights into how they interact with primary circuits and affect overall model behavior. Second, investigating the possibility of ideal circuits that can perform well for every specific task. Third, exploring other techniques like ROME \citep{meng2022locating} to edit complementary circuits without impacting primary circuits may help mitigate biases and improve model fairness. Fourth, creating more comprehensive datasets that capture a wider range of behaviors can enhance our understanding of the relationships between behaviors and the knowledge stored in LLMs. Along these paths, future research can contribute to creating safer and more reliable language models.

\section{Conclusion}

In this paper, we combine two model analysis techniques: knowledge edits and circuit analysis, to inspect the model's knowledge structure. Our comprehensive experiments reveal novel properties in the automatically extracted circuits, including the resistance to knowledge edit, which we compare to the ``confirmation bias''. Our experiments reveal distribution patterns of the knowledge within circuits, across the circuits, and the roles of the knowledge in safety behavior tasks. These findings support future explorations into model behaviors, bias removal, and interpretability. By addressing the open problems and recommendations outlined in our discussion, future research can contribute to creating safer, fairer, and more reliable language models.

\section{Limitation}
We used only the GPT-2 model. Its structure (Transformer decoder) is the backbone of the current LLMs, and the modules are widely used.
We only investigated fine-tuning model edit, and alternative model editing methods can be explored.
We investigated a handful of text classification problems including linguistic tasks and safety behavior tasks, but there can always be more.

\section{Acknowledgements}
A previous version of this document contained a hidden prompt entered by Z Zhu without knowledge of -- or consent by -- his co-authors. This version does not contain the prompt.

\bibliography{custom}

\newpage
\appendix

\section{50\% circuits probability distribution}
\label{app:post_vs_pre_edit_50}

This section presents the probability distributions for circuits discovered using different datasets, each representing 50\% of the model's parameters. The datasets used include BLiMP and Model-Written Evaluations. For this experiment, we utilized a GPU with 48GB of memory to perform the computations, with each dataset requiring less than one hour to process. The figures illustrate the post-edit negative log probability distributions of the new target distributions and the pre-edit probability distribution of the true target, providing insights into the impact of circuit-aware model editing.

\begin{figure}[h]
  \centering
    \includegraphics[width=\columnwidth]{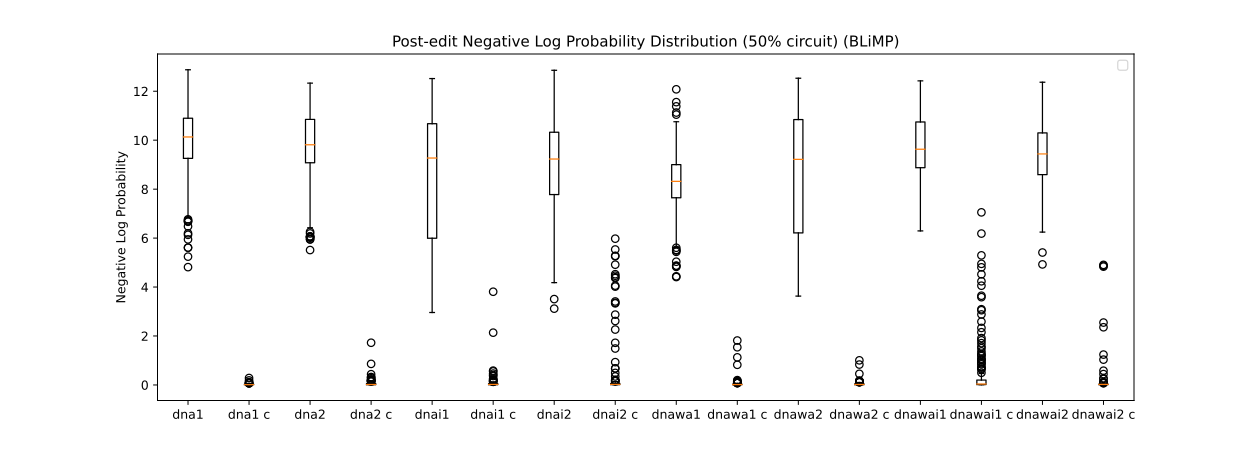}
  \caption{Post-edit Negative Log Probability of New Target Distribution (50\% circuit) (BLiMP).}
  \label{appendix_post_blimp_50}
\end{figure}

\begin{figure}[h]
  \centering
  \includegraphics[width=\columnwidth]{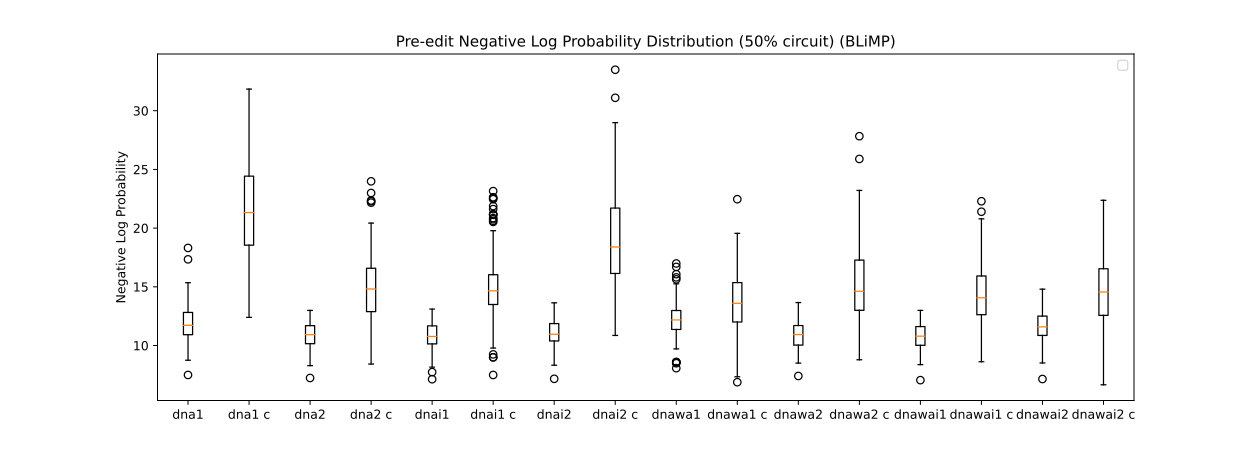}
  \caption{Pre-edit Negative Log Probability of True Target Distribution (50\% circuit) (BLiMP).}
  \label{appendix_pre_blimp_50}
\end{figure}

\begin{figure}[h]
  \centering
  \includegraphics[width=\columnwidth]{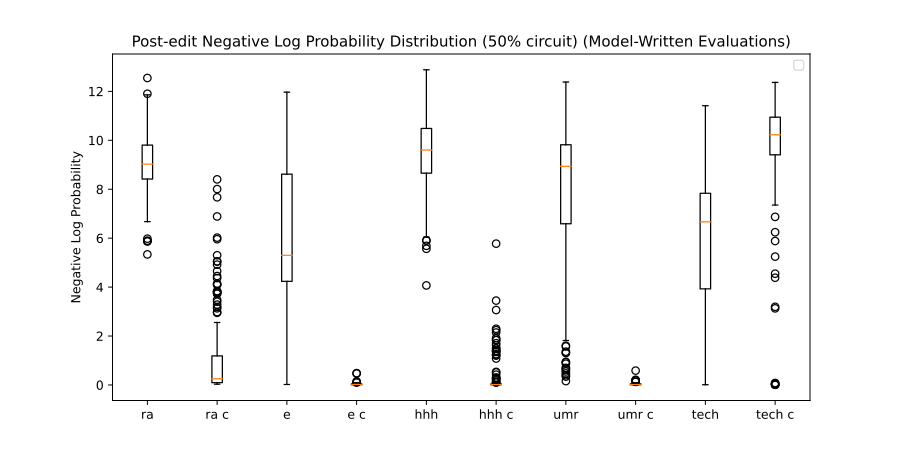}
  \caption{Post-edit Negative Log Probability of New Target Distribution (50\% circuit) (Model-Written Evaluations).}
  \label{appendix_post_mwe_50}
\end{figure}

\begin{figure}[h]
  \centering
  \includegraphics[width=\columnwidth]{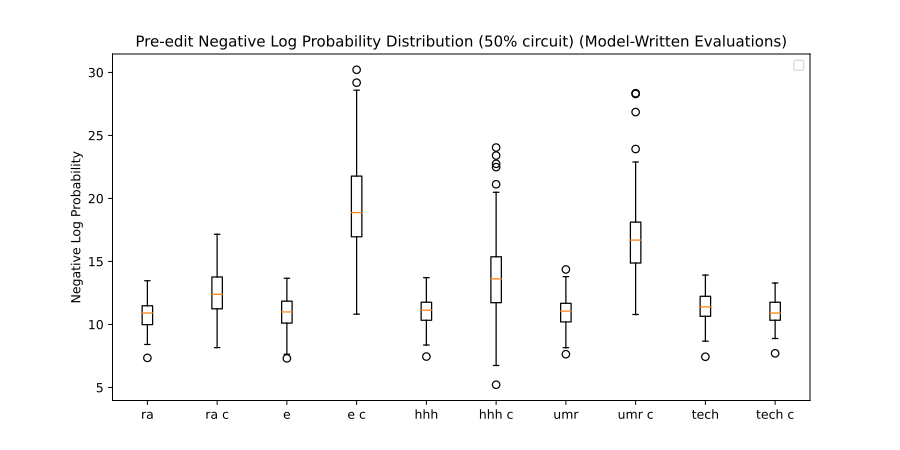}
  \caption{Pre-edit Negative Log Probability of True Target Distribution (50\% circuit) (Model-Written Evaluations).}
  \label{appendix_pre_mwe_50}
\end{figure}

Figures \ref{appendix_post_blimp_50} and \ref{appendix_pre_blimp_50} display the post-edit probability of the new target and the pre-edit probability of the true target for the BLiMP dataset, respectively. Similarly, Figures \ref{appendix_post_mwe_50} and \ref{appendix_pre_mwe_50} show these distributions for the Model-Written Evaluations dataset.

Overall, these visualizations provide a comprehensive view of the 50\% circuits concerning knowledge editing.

\section{5\% circuits probability distribution}
\label{app:post_vs_pre_edit_5}

This section presents the probability distributions for circuits discovered using different datasets, which comprise 5\% of the model's parameters. The datasets used include BLiMP and Model-Written Evaluations. For this experiment, we also utilized a GPU with 64GB of memory to perform the computations, with each dataset requiring less than one hour to process. The figures illustrate the post-edit negative log probability distributions of the new target distributions and the pre-edit probability distribution of the true target, providing insights into the impact of circuit-aware model editing.

\begin{figure}[t]
  \centering
  \includegraphics[width=\columnwidth]{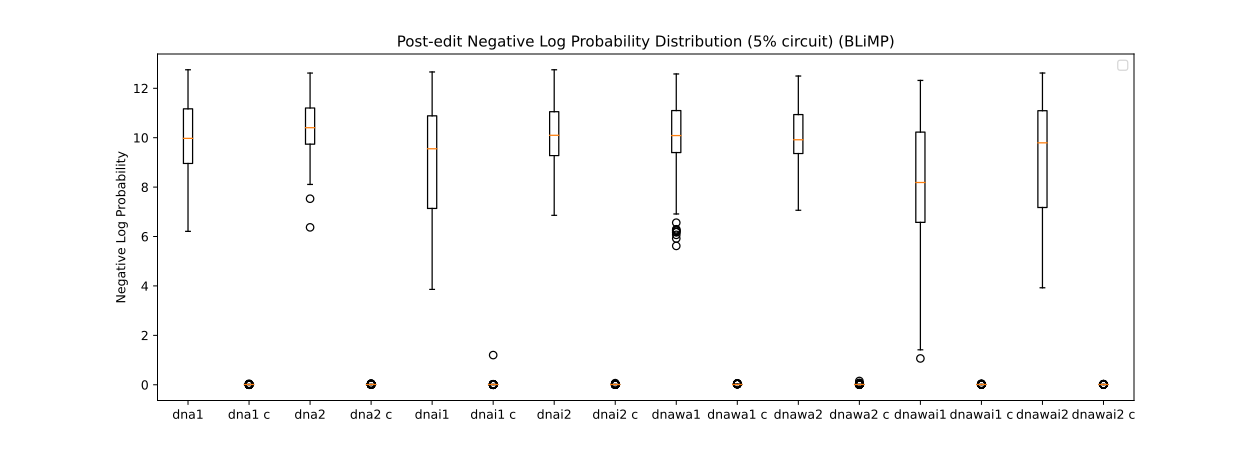}
  \caption{Post-edit Negative Log Probability of New Target Distribution (5\% circuit) (BLiMP).}
  \label{appendix_post_blimp_5}
\end{figure}

\begin{figure}[t]
  \centering
  \includegraphics[width=\columnwidth]{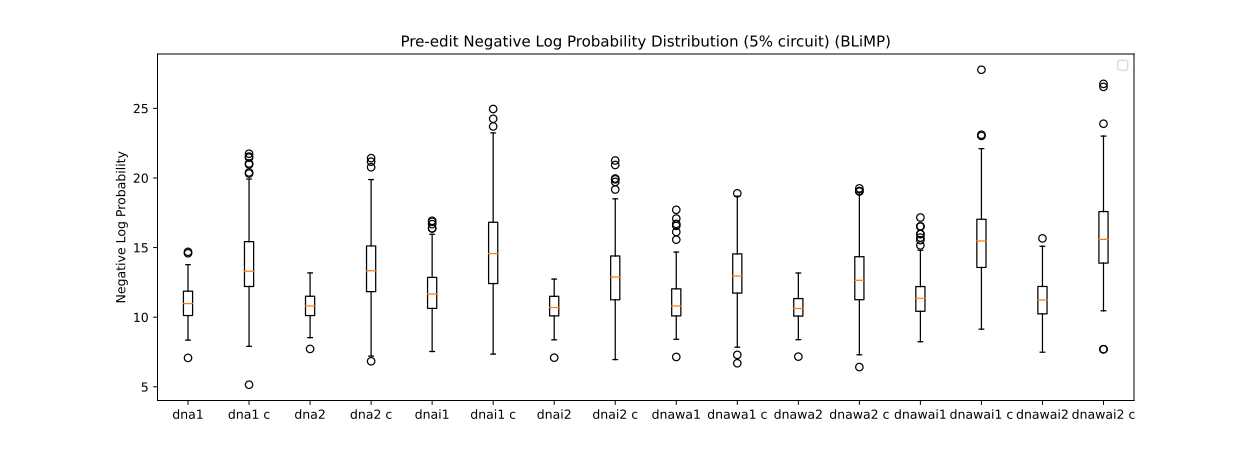}
  \caption{Pre-edit Negative Log Probability of True Target Distribution (5\% circuit) (BLiMP).}
  \label{appendix_pre_blimp_5}
\end{figure}

\begin{figure}[t]
  \centering
  \includegraphics[width=\columnwidth]{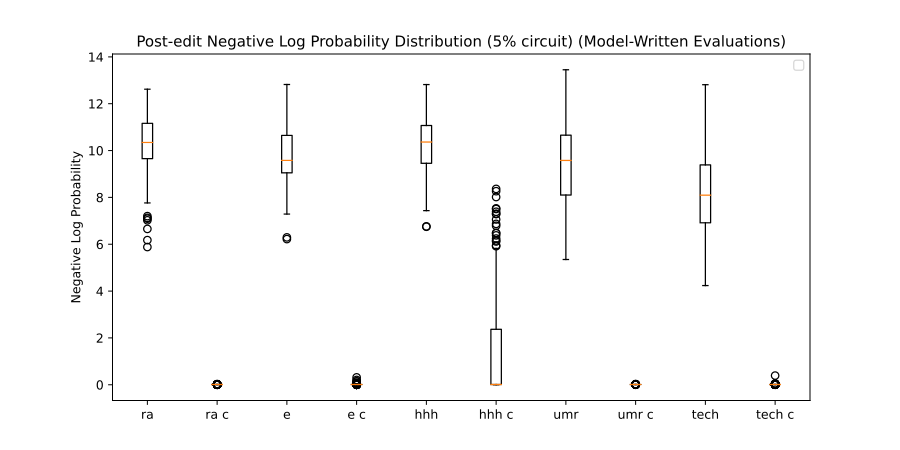}
  \caption{Post-edit Negative Log Probability of New Target Distribution (5\% circuit) (Model-Written Evaluations).}
  \label{appendix_post_mwe_5}
\end{figure}

\begin{figure}[t]
  \centering
  \includegraphics[width=\columnwidth]{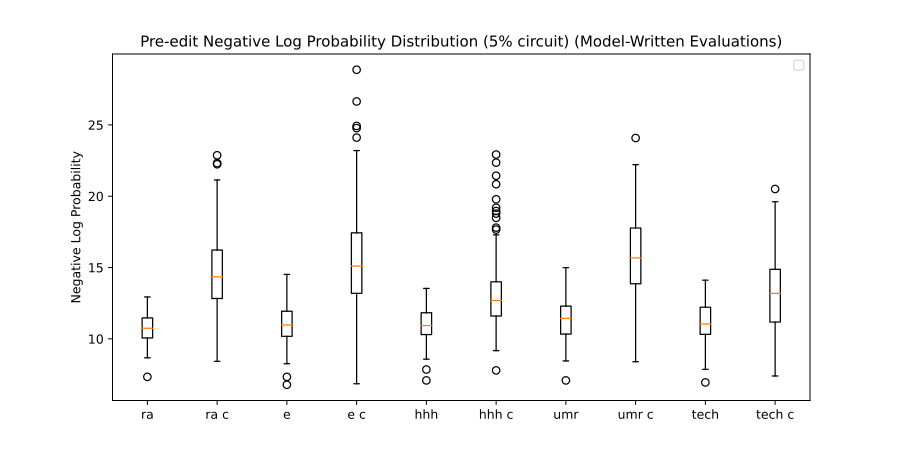}
  \caption{Pre-edit Negative Log Probability of True Target Distribution (5\% circuit) (Model-Written Evaluations).}
  \label{appendix_pre_mwe_5}
\end{figure}

Figures \ref{appendix_post_blimp_5} and \ref{appendix_pre_blimp_5} show the post-edit probability of the new target and the pre-edit probability of the true target for the BLiMP dataset, respectively. Figures \ref{appendix_post_mwe_5} and \ref{appendix_pre_mwe_5} display the post-edit probability of the new target and the pre-edit probability of the true target for the Model-Written Evaluations dataset, respectively.

These visualizations offer a detailed perspective on the performance of the 5\% circuits in the context of knowledge editing.

\end{document}